# Investigation of UAV Detection in Images with Complex Backgrounds and Rainy Artifacts

Adnan Munir[2], Abdul Jabbar Siddiqui[1,2], Saeed Anwar[1,3] [1]
SDAIA-KFUPM Joint Research Center for Artificial Intelligence [2] [3]
Computer Engineering Department; Information and Computer Science Department
King Fahd University of Petroleum and Minerals, Dhahran, Saudi Arabia
{g202110550, abduljabbar.siddiqui, saeed.anwar}@kfupm.edu.sa

## Abstract

*To detect unmanned aerial vehicles (UAVs) in real-time, computer vision and deep learning approaches are evolving research areas. Interest in this problem has grown due to concerns regarding the possible hazards and misuse of employing UAVs in many applications. These include potential privacy violations. To address the concerns, visionbased object detection methods have been developed for UAV detection. However, UAV detection in images with complex backgrounds and weather artifacts like rain has yet to be reasonably studied. Hence, for this purpose, we prepared two training datasets. The first dataset has the sky as its background and is called the Sky Background Dataset (SBD). The second training dataset has more complex scenes (with diverse backgrounds) and is named the Complex Background Dataset (CBD). Additionally, two test sets were prepared: one containing clear images and the other with images with three rain artifacts, named the Rainy Test Set (RTS). This work also focuses on benchmarking state-of-the-art object detection models, and to the best of our knowledge, it is the first to investigate the performance of recent and popular vision-based object detection methods for UAV detection under challenging conditions such as complex backgrounds, varying UAV sizes, and low-toheavy rainy conditions. The findings presented in the paper shall help provide insights concerning the performance of the selected models for UAV detection under challenging conditions and pave the way to develop more robust UAV detection methods. The codes and datasets are available at: https://github.com/AdnanMunir294/UAVDCBRA.*

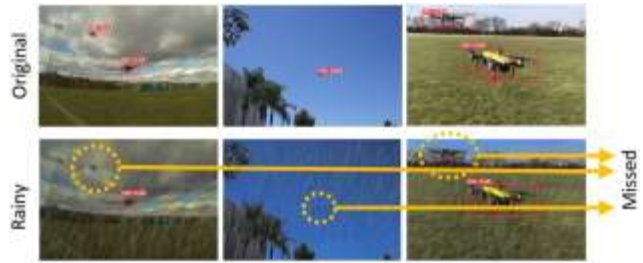

Figure 1. Rain artifacts degrade the performance of a state-of-theart model for UAV detection; bottom-row shows the missed UAVs and low-confidence detections of the model due to rain artifacts. These vividly highlight the adverse effects of weather and complex backgrounds on UAV detection accuracy, highlighting the implications of rain-induced distortions and artifacts for detection and localization models in real-world surveillance.

## 1. Introduction

UAVs have been gaining increasing popularity in recent years for a variety of uses, including aerial photography, delivery services, and surveillance. However, the proliferation of drones has generated privacy, security, and safety issues. Drones can be employed for harmful objectives, such as carrying illegal goods or performing uninvited monitoring. There is an increasing demand for effective drone detection and identification technologies to address these security and privacy concerns.

Vision-based analysis (of images and videos), which uses computer vision algorithms to detect and track drones in real-time, is one of the most promising drone detection approaches. Such vision-based systems have advantages over other approaches (e.g., radar-based), including improved accuracy, cheaper costs, and the potential to gather visual evidence for prosecution [1]. Several vision-based UAV/drone detection methods have been presented in the literature based on object detection methods, motion analysis, and machine learning-based algorithms. Object detection methods recognize the UAVs/drones as objects of interest in photos or videos, whereas motion analysis techniques detect the drones' movement patterns. Machine learning-based algorithms use methods such as convolutional neural networks (CNNs) to train the system to detect the visual aspects of the drone [2].

The methods used to detect, identify, and locate unmanned aerial vehicles (UAVs) in specific airspace are referred to as UAV or drone detection methods. To detect and track drones, several sensors such as radar, electrooptical/infrared (EO/IR) cameras, and acoustic sensors can be used. Radar systems detect drones by detecting radio wave reflections, whereas acoustic sensors detect the specific sound signature of UAVs. Drones may be detected using EO/IR cameras that recognize visual features such as form, size, and color [3]. Furthermore, some drone detection systems analyze communication data and discover patterns related to drone activity using artificial intelligence and machine learning algorithms. These different sensing modalities present their respective pros and cons. This work focuses on vision-based UAV detection methods.

The UAV detection methods in the literature have not been well investigated for multi-scale UAV detection, especially under the effect of weather conditions such as rain. This motivates us to study how cutting-edge object detection methods would perform for detecting UAVs of varying scales in varying complex backgrounds, especially under the effect of rain.

Contributions: This paper makes the following contributions:

- Two types of datasets with UAVs and Birds are collected and curated in this study: one with only a sky background and the other with diverse, complex

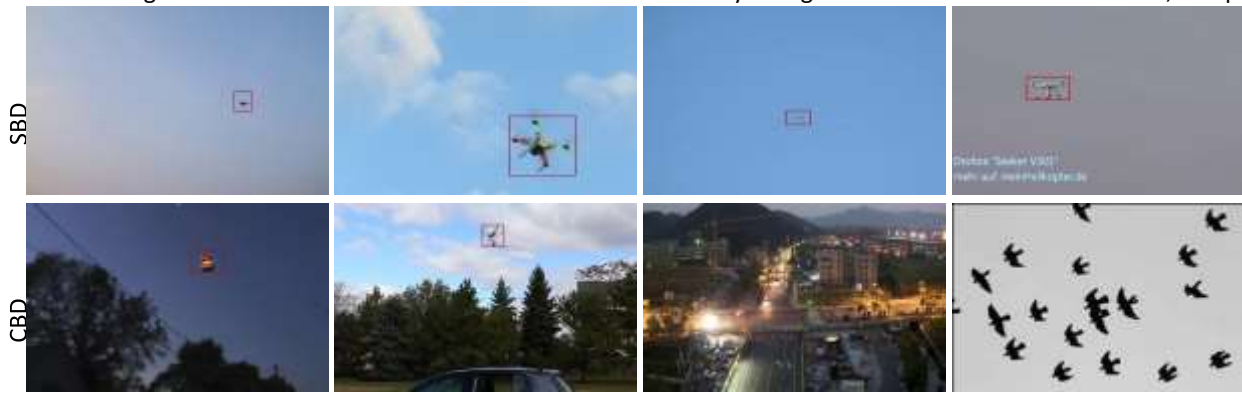

Figure 2. Sample images from the proposed Sky Background Dataset (SBD) and Complex Background Dataset (CBD) are shown. The red bounding boxes show the original location of the UAV in each image.

  environments such as urban, semi-urban, indoor, outdoor, mountains, sunny, and cloudy, each with various lighting conditions.

- To the best of our knowledge, this is the first work to investigate the performance of one-stage and twostage object detectors for UAV detection on the proposed datasets.

- In addition, for the first time to the best of our knowledge, this work investigates and analyses the impact of three types of rain artifacts on the UAV detection performance of object detectors on a novel test set. The datasets will be publicly released for reproducibility and further research.

## 2. Related Work

This work focuses on vision-based UAV detection; hence, this section reviews relevant related works.

In [4], the authors investigated the recognition of small drones in videos using deep learning techniques. Authors experimented with CNN models such as ResNet-101, Inception with Faster RCNN, and Single Shot Detector (SSD) [5]. The tests that combined the ResNet-101 basic architecture with Faster-RCNN produced the best results. The authors trained the network with transfer learning from publicly accessible pre-trained COCO models to hasten convergence using the Drone vs Bird dataset [6] in their work. The Faster RCNN with ResNet101 achieves better mean average precision (mAP) than InceptionV2 [7] and SSD. As they are comparing the different algorithms, the execution time and statistical testing of each model are not performed.

The work of [8] proposed a drone detection system based on the Birds vs Drone dataset [6]. The model is then evaluated using their custom-collected dataset from two different drone videos. YOLOv4 [9] model is fine-tuned on the custom dataset. By using transfer learning, they made the Darknet [10] framework compatible with their own proposed system. The last three YOLO and convolutional layers were fine-tuned on the two classes. The custom dataset used in their study is collected from different drone videos with additional sunlight and background conditions. The achieved video detection frame per second (FPS) is 20.5. Only a custom dataset is used to tune the neural network. Their work is only limited to YOLOv4 [9], and no other object detector is examined.

In [11], with the help of 241 videos containing 331,486 images, the authors applied and tested four detection and three tracking algorithms. The detection algorithm achieves an overall reasonable mAP, while the tracking algorithm obtains a good Multi-Object Tracking Accuracy (MOTA). The authors considered the Faster-RCNN [12], YOLOv3 [13], SSD [5], and DETR [14] methods for object detection. The SORT [15] and DeepSORT [16] algorithms are utilized for object tracking. This study combined the MAVVID [17], Anti-UAV [18], and Drones-vs-Bird datasets for training and testing. However, none of the used datasets included adverse weather conditions. Although Faster RCNN is reported to achieve the highest mAP for recognizing tiny UAVs, YOLOv3 [19] had the highest overall mAP. DETR [14] works effectively with cross-modal videos as a detection backbone for tracking systems for microscopic objects.

A multi-featured and advanced UAV detection network for SafeSpace is proposed by [20], based on an improved version of the YOLOv5 [21] detection method. For accurate and fast detection of small objects, authors changed the backbone and neck of the YOLOv5 network to develop MFNet. To increase the sensitivity and scalability of feature extraction in the YOLOv5s model, the authors changed the kernel size (KS) in the backbone and neck, resulting in changes to the size of the extracted feature maps. Three versions of MFNet were proposed: MFNet-S, MFNet-M, and MFNet-L, based on their small, medium, and large kernel and feature map sizes. Authors collected over 5105 images of UAVs and birds from the publicly accessible open-source datasets on the Roboflow [22] to train the proposed MFNet architecture.

Compared to the YOLOv5s model, the MFNet-M model achieved a better average precision, average recall, mAP, and IoU, which also gets substantial benefits for other MFNet models. MFNet-M obtained the best precision on the Birds class; however, for UAVs, the maximum precision is also reasonable and better than other versions of MFNet. However, their work only considered birds and one type of drone, which makes their developed models suffer in cases of varying drone types. For example, kites in the background were also detected as drones by their method.

## 3. Proposed Datasets

As a part of this work, several datasets were reviewed and acquired from various publicly accessible or published sources. Following the review, several datasets were merged to build custom and more challenging datasets with two distinct categories for training and one for testing. The first category, named *Sky Background Dataset*

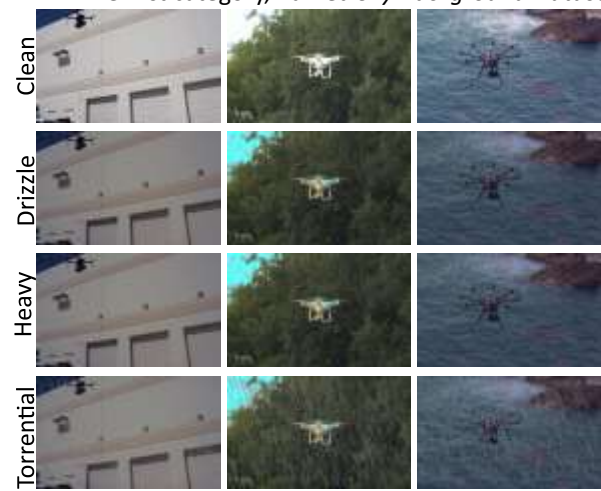

Figure 3. Sample images from the proposed Complex Background Test Set (CBTS) and Rainy Test Set (RTS). Three types of rain artifacts applied are *drizzle, heavy* and *torrential*. The top row shows the original clean image (before applying artifacts).

*(SBD)*, comprises 3888 images, which exclusively feature a sky background, as illustrated in Fig. 2 (top-row). The second category with 12,492 images, named *Complex Background Dataset (CBD)*, includes diverse backgrounds such as urban, semi-urban, indoor, outdoor, mountains, sunny, and cloudy, each with different lighting conditions, as depicted in Fig. 2 (bottom-row). The third dataset, named the *Rainy Test Set (RTS)*, is composed of images with three types of rainy artifacts (see Fig. 3). A summary for all the datasets introduced in this article is presented in Table 1. Sky Background Dataset (SBD): The SBD is formed by merging three different datasets (selecting images with the only sky as background), namely: Det-Fly [23], Wosdec challenge [24], and a real-world object detection dataset designed for quadcopters [25].

Complex Background dataset (CBD): While reviewing other datasets, some limitations were observed. For example, the MIDGARD [26], Det-Fly [23], and Anti-UAV datasets did not include the birds in their datasets. Some of the datasets only contain a single UAV model. The CBD is a combination of seven datasets (selecting only images with complex backgrounds), including Det-Fly [23], Wosdec challenge [24], real-world object detection dataset for quadcopters [25], MIDGARD [26], Real-Time Drone Detection and Tracking [27], and Vision-based Anti-UAV [28]. Annotations from various formats were transformed into YOLO and JSON formats. To make the datasets more challenging and to investigate the ability of detectors to differentiate between UAVs and birds, images of birds flying were added to the dataset.

Table 1. A summary of all the novel datasets used for training, validation and testing.

| Datasets | SBD | CBD | Rainy Test Set | | |
| --- | --- | --- | --- | --- | --- |
| | | | Drizzle | Heavy | Torrential |
| Training | 3456 | 9725 | - | - | - |
| Validation | 432 | 2767 | - | - | - |
| Testing | - | - | 2528 | 2528 | 2528 |

Rainy Test Set (RTS): Many of the prior works on vision-based UAV detection focused on clear images, e.g., in the Anti-UAV [28] dataset, rainy images were not considered. In this work, RTS is built based on the Anti-UAV [28] test set, hereby referred to as the Test Set (TS), which is not included in the training/validation data of SBD and CBD datasets.

Rainy conditions could cause performance degradation of the UAV detection methods, a matter not studied in prior works to the best of our knowledge. Rain is composed of innumerable drops of various sizes and complicated forms, and it spreads very randomly and at varying rates when it falls on streets, pavement, automobiles, pedestrians, and UAVs in the scene. Raindrops can cause changes in the pixel values of images and video frames. This is because the drops block some of the reflected light from the objects in the scene. Additionally, rain streaks can make the visual data appear less contrasting and whiter. To consider these challenging conditions, rain streaks were synthetically added to the test images of the Anti-UAV [28] dataset to build the Rainy Test Set (RTS) proposed in this work. The clean images of RTS are utilized to obtain baseline performance of selected models, while the rainy images of RTS are used to analyze the impact of rainy artifacts.

Three types of rain effects (drizzle, heavy, and torrential) were added to the TS dataset images and generated rainy test set drizzle ($RTS_{drizzle}$), rainy test set heavy ($RTS_{heavy}$), and rainy test set torrential ($RTS_{tor}$). A popular tool employed to generate synthetic (yet realistic) rain effects on images is the automold library of [29] as used in the literature of other domains [30–32]. It is worth mentioning that the rain effects are added only to the test set images (and not to the training sets) to study the performance of the selected models under investigation. While one may argue that including rain effects in training images could enhance a model's accuracy on rainy test images, this paper chooses to train the selected models only on clean (no rain) data to facilitate a better understanding of the strengths and weaknesses of the preferred models in unforeseen rainy conditions.

## 4. Methods for Benchmarking

This work seeks to investigate the impact of complex backgrounds and rainy weather conditions on the performance of UAV detection methods and examine the advantages or disadvantages of the selected methods. To this end, a custom test dataset is curated (i.e., the RTS described in the previous section), considering challenging conditions such as varying UAV scales, background variety, and rain effects. The object detection methods selected to be investigated in this work include critical one-stage detectors such as YOLOv5 [21], YOLOv8 [33], RetineNet [34], and a popular two-stage detector called Faster-RCNN [35]. In the following subsections, a brief overview and description of the selected object detectors are provided.

YOLOv5 [21]: The model consists of four main components: input, Darknet-53 [13] backbone network, PANet [36] neck network, and output. The basic structures of YOLOv5 are CBS and $CSP_n$. CBS is a basic convolution module that includes a batch normalization operation and a SiLU activation function [37] to prevent gradient disappearance. $CSP_n$ consists of two branches; the first is a series of n Bottleneck modules, while the second is a CBS convolution block. These two branches are stacked to increase the network depth and enhance feature extraction capabilities. The input terminal is responsible for improving data and adapting anchor boxes. The backbone network comprises 2 $CSP_1$ and 2 $CSP_3$ structures for extracting features. By using multiple down-samplings and up-samplings, the neck network combines features of different levels. The output head performs bounding box regression and NMS (Non Maximal suppression) [38] post-processing to achieve precise target detection.

YOLOv8 [33]: is an enhanced version of YOLOv5 produced by [39]. The Ultralytics YOLOv8 is the most recent variant of the YOLO-based models used for object detection and image segmentation. This advanced model improves on its predecessors by adding new features and enhancements that boost its performance, adaptability, and effectiveness [39]. The YOLOv8 model surpasses its earlier versions by integrating a novel backbone network, an anchor-free split head, and updated loss functions. These upgrades in YOLOv8 were reported to produce superior outcomes. The anchor-free approach aims to predict the center of an object of interest directly rather than the offsets from a known anchor box. The NMS [38] is accelerated by anchor-free detection since it minimizes the number of box predictions. In YOLOv8, the first 6×6 convolution in the backbone is replaced with a 3×3 convolution block, and two of the convolutions (No.10 and No.14 ) were removed from the YOLOv5 setup.

RetinaNet [34]: is an object detection network that includes a Feature Pyramid Network (FPN) [40], a backbone network, and two classification and regression subnetworks. The backbone network is in charge of extracting features from the input images, and it employs ResNet [41] to generate four feature maps with varying resolutions. The FPN then combines these feature maps to form a pyramid of multi-scale feature maps, which are utilized to identify objects of varied sizes. RetinaNet is intended to manage unbalanced data and objects of various sizes, which it does through its unique architecture and usage of the Focal Loss function [42]. It is a one-stage object detection model [34]. RetinaNet addresses the difficulties of unbalanced data and objects of various sizes through a special design that uses the Focal Loss function and the Feature Pyramid Network (FPN) [40].

Faster-RCNN [12]: Fast R-CNN [43] is a two-stage object detection network that overcomes various shortcomings of the prior R-CNN network [44] by increasing speed. The ROI Pooling layer in Fast R-CNN gathers feature vectors of the same length from each Region of Interest (ROI) inside an image. Faster R-CNN goes one step further by employing a single-stage network rather than R-CNN's three stage method. It computes convolutional layer calculations once and then distributes them across all suggestions (such as ROIs). Faster R-CNN outperformed R-CNN in terms of accuracy [35].

## 5. Experiments and Results

The performance of selected UAV detection methods will be investigated in the following settings: (i) on SBD images (no complex backgrounds, no rain), (ii) on CBD images (with complex backgrounds but no rain), and (iii) on Rainy Test Set (RTS) images.

Evaluation Metrics: The performance will be evaluated based on the mean average precision ($mAP_{50}$), mean average precision ($mAP_{50-95}$), which are mostly cited and adopted by the Computer Vision Society for comparing the performance of object detection models. Moreover, we provide insights via GradCAM into the models' performance. The experimental results in this paper present the comparative evaluation and analysis of well-known one-stage and two-stage object detectors for the proposed custom dataset.

The SBD and CBD datasets are each divided into training and validation sets with a ratio of 80% and 20%, respectively. The TS dataset is utilized for testing the models. The Google Colab [45] is employed for training the models using NVIDIA A100 [46] GPUs for 100 epochs. The FasterRCNN and RetinaNet are trained using Detectron2 [47] library. Faster-RCNN and RetinaNet have Resnet101 as a backbone with a Feature pyramid network (FPN), while YOLOv8 and YOLOv5 [21] utilized Darknet-53 to extract features from images.

UAV Detection in Sky Images (SBD): The SBD is used to train and validate one-stage object detectors such as Yolov5, Yolov8, RetinaNet, and two-stage object detectors FasterRCNN. The mAP results are presented in Table 2. The results indicate a better performance of Yolov8 than Yolov5 and other models. The Yolov8 can perform with high speed in a real-time environment. The Yolov8 has higher training parameters than Yolov5, so it takes longer to train than UAV Detection in Complex Backgrounds (CBD): This section describes the validation and testing results of the models studied in this work, including YOLOv8, YOLOv5, Faster-RCNN, and RetinaNet. Table 3 presents the different models' performance on CBD. The YOLOv5 performs the best on the testing set in terms of both $mAP_{50}$ and $mAP_{50-95}$. RetinaNet showed a bit improved performance than Faster-RCNN. However, Faster-RCNN's mAP could have improved with an increased number of epochs (> 100); regardless, in this study, we seek to examine and compare the different models' performance, each restricted to 100 training epochs. UAV detection outputs of the selected models on the TS images are shown in Fig. 4.

Table 2. An overview of the performance of various object detection models for UAV detection on the Sky Background Dataset's (SBD) Validation Set. The boldface numbers show the best performance.

| Model | Backbone | $mAP_{50}$ | $mAP_{50-95}$ | Inference Time (ms) |
| --- | --- | --- | --- | --- |
| YOLOv5m | Darknet53 | 87.7 | 71.7 | 6.2 |
| YOLOv8m | Darknet53 | 88.3 | 72 | 7.5 |
| Faster-RCNN | ResNet-101 + FPN | 86.53 | 68.1 | 41.3 |
| RetinaNet | ResNet-101 + FPN | 84.04 | 63.3 | 29.2 |

Table 3. The performance comparison of various object detection models for UAV detection on the CBD's Validation Set and on the Test Set (TS) for testing. YOLOv5m shows the best performance among others.

| Model | Backbone | Validation | | Test | |
|---|---|---|---|---|---|
| | | $mAP_{50}$ | $mAP_{50-95}$ | $mAP_{50}$ | $mAP_{50-95}$ |
| YOLOv5m | Darknet-53 | 95.6 | 57.5 | 94.6 | 59 |
| YOLOv8m | Darknet-53 | 91.3 | 58.1 | 91.8 | 51 |
| Faster-RCNN | ResNet-101 + FPN | 78.04 | 41.34 | 82 | 46 |
| RetinaNet | ResNet-101 + FPN | 80.85 | 44.39 | 84.39 | 46.75 |

Figure 4. Sample UAV detection outputs of the selected models on TS images with complex backgrounds. False Positives of RetinaNet and of Faster-RCNN are shown on (a) and (c), respectively, which demonstrate the low performance of these models compared to that of Yolov5. The $AP$ of models may be improved by increasing the number of images in SBD.

UAV Detection in Rainy Conditions: To study the effect of rain on the UAV detection performance, the rainy test sets: drizzle ($RTS_{drizzle}$), heavy ($RTS_{heavy}$), and torrential ($RTS_{tor}$) are used. The evident degradation in mAP and IoU scores is observed in the rainy images.

Fig. 5 shows four sample images with and without rain effects. In Fig. 5(a), and Fig. 5(d), RetinaNet and FasterRCNN are able to detect the UAV in the original test image and with rain artifacts but with lower confidence. Similarly,
YOLOv5 (best viewed in color, zoomed in).

in Fig. 5(b), and Fig. 5(c), the YOLOV5 and YOLOv8 are able to detect UAV in original images with higher confidence; however, they fail to detect in the image with rain artifacts. We evaluate the selected models on the three rainy test sets. Table 4 shows the mAP on rainy test images and the inference speed of each model. The overall performance degrades due to rain artifacts in the test images. The performance degradation of YOLOv5m, YOLOv8m, FasterRCNN, and RetinaNet is 50.62, 53.23, 56.25, and 58.40 percentage points, respectively. The YOLOv5 model performance degradation is shown in Fig. 1. The RetinaNet shows the worst percentage of performance decline in mAP. Table 4 shows that the degradation in mAP increases with the amount of rain.

Performance Analysis: To develop an effective detection system, the GradCAM (Gradient-weighted Class Activation Mapping) [48] is often employed by researchers to scrutinize a model's focus areas and understand its outputs. To get more insights into each model's outputs, we integrate GradCAM with all four models. The GradCAM highlights the region that influences the model's prediction. We chose images from the heavy-rain test dataset to apply GradCAM to assess the performance of the four mentioned models for UAV detection. For example, the GradCAM outputs on a sample image with and without rain effects are shown in Fig. 6. The top row shows the respective model's attention towards the UAV, and all four models detected the UAV in the image.

Observing the bottom row of Fig. 6 (with rain artifacts), the Faster-RCNN and the YOLOv5 model's attention is towards the UAV, which explains why the UAV's presence is detected in the image. On the contrary, the GradCAM maps for RetinaNet and YOLOv8 on the rainy image (bottom-

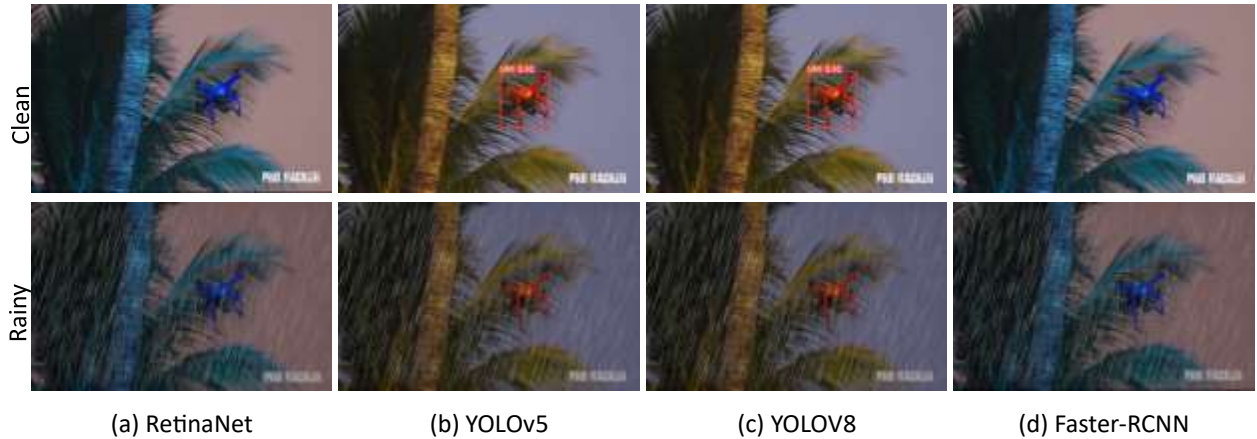

(a) RetinaNet  (b) YOLOv5  (c) YOLOV8  (d) Faster-RCNN

Figure 5. Comparisons on clean TS images and rainy RTS images. RetinaNet and Faster-RCNN detect the UAV in the rainy image with lower confidence; however, YOLOv5 and YOLOv8 do not detect the UAV in the rainy image (best viewed on screen when zoomed-in).

Table 4. Summary of the performance of various object detection models for UAV detection on the three Rainy Test Sets (RTS): $RTS_{drizzle}$, $RTS_{heavy}$ and $RTS_{tor}$ vs. the performance on clean TS images (no rain).

| Model | Backbone | mAP$_{50}$ | | | | Inference time (ms) (%) |
|---|---|---|---|---|---|---|
| | | no rain (%) | drizzle (%) | heavy (%) | torrential (%) | |
| YOLOv5m | Darknet-53 | 95.6 | 66.8 | 65.1 | 47.2 | 6.2 |
| YOLOv8m | Darknet-53 | 91.3 | 65.7 | 60.2 | 42.7 | 7.5 |
| Faster-RCNN | Resnet-101 + FPN | 85.5 | 50.35 | 45.25 | 37.40 | 41.3 |
| RetinaNet | Resnet-101 + FPN | 82.5 | 46.33 | 43.69 | 35.57 | 29.2 |

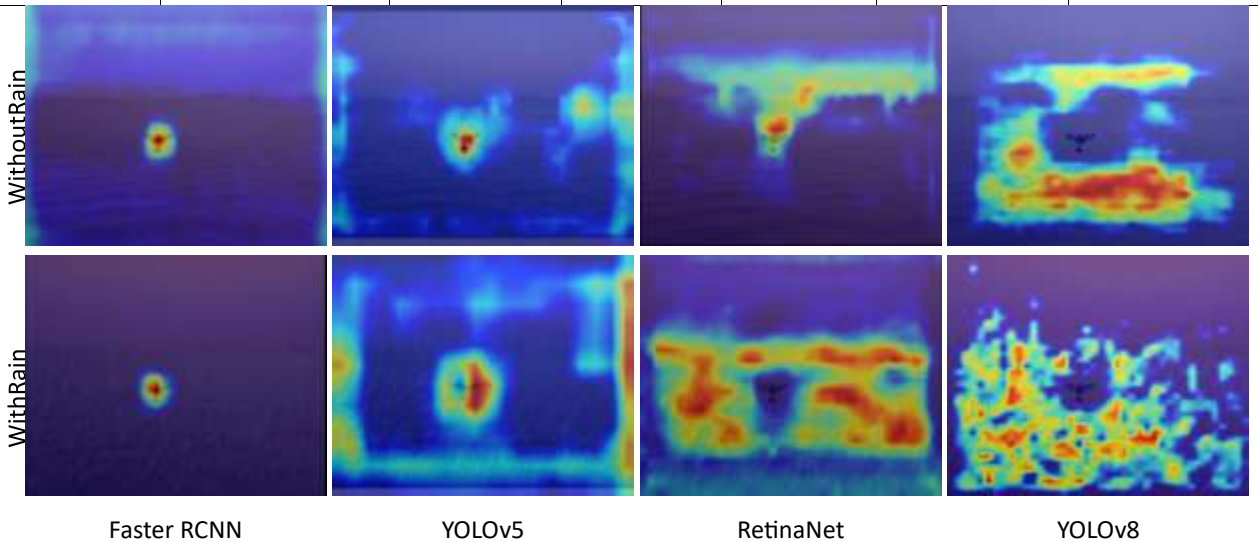

Faster RCNN  YOLOv5  RetinaNet  YOLOv8

Figure 6. Sample GradCAM results are shown with and without rain artifacts. The Faster RCNN and YOLOv5 show high scores for the UAV class, while RetinaNet and YOLOv8 show lower class scores.

row) show that the models are attending to a region with no UAV, which misled the models and caused a miss detection. In fact, the GradCAM maps explain how the affected model focuses on regions scattered around that do not have any UAVs.

These observations demonstrate how Faster-RCNN and YOLOv5 can detect the UAVs in the image despite the rainy conditions. However, the RetinaNet and YOLOv8 struggled to detect the UAV in the rainy image and demonstrated lower robustness. The relatively robust performance of

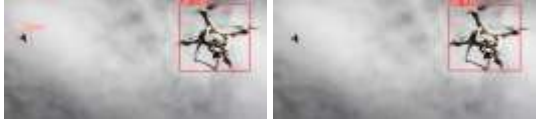

Faster-RCNN despite the rainy artifacts could be attributed to the two-stage architecture of Faster-RCNN, which has a dedicated region proposal network to keep the model focused on the regions of interest.

Failures Detection: Missed detections may arise due to various factors, including low image quality, object occlusion, insufficient model training, or insufficient data. In applications related to anti-UAV, missed detection could lead to security threats. Fig. 7 shows the detection outputs of the selected models for a sample image from the CBD. The Faster-RCNN and YOLOv5 detect the UAV with high confidence. RetinaNet and YOLOv8 fail to detect the UAV in the same image. Faster-RCNN fails to detect the UAV after applying rain effects on the same image, as shown in Fig. 7 (d). The missed detection rate on the TS is found to be 2%, 5%, 3%, and 1% for FasterRCNN, RetinaNet, YOLOv8, and YOLOv5, respectively.

Detection of UAVs vs Birds: Distinguishing between birds and UAVs is one of the difficult challenges for real-world UAV detection methods. The complex dataset is curated with many bird samples in its training set. The WOSDETC [24] "Birds vs Drone" dataset is included in our CBD. For the test set, images are taken from the internet [49]. YOLOv5 and YOLOv8 are trained on the CBD, which included bird images. However, the YOLOv8 model failed to detect the bird in the test image shown in Fig. 8. While YOLOv5 detects both birds and drones with a higher confidence score. Table 5 presents the average precision ($AP_{50-95}$) of each model for UAV and Bird categories. Specifically, for the Bird class, Faster-RCNN shows poor performance as compared to the other three models. On the other hand, RetinaNet gets low AP for UAV class.

Table 5. UAV vs Bird detection performance of selected models.

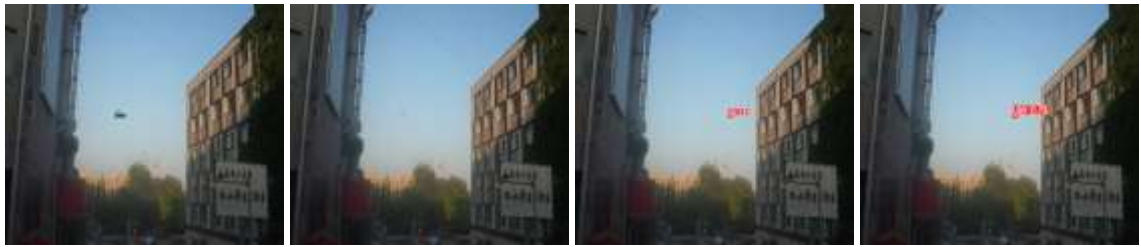

| | Faster-RCNN | RetinaNet | YOLOv5 | YOLOV8 |

Figure 7. Sample detection outputs: Faster-RCNN, Yolov8, and YOLOv5 detect the UAV; however, RetinaNet fails to detect the UAV. Figure 8. YOLOv5 and YOLOV8 detect the UAV with high confidence; YOLOv8 fails to detect bird (best viewed zoomed in).

| Model | $AP_{UAV}$ (%) | $AP_{Bird}$ (%) |
|---|---|---|
| YOLOv5 | 69.3 | 48.6 |
| YOLOv8 | 67.2 | 48.5 |
| Faster-RCNN | 59.2 | 29.3 |
| Retina-Net | 57.7 | 35.9 |

## 6. Conclusion

In this paper, comprehensive evaluation experiments were conducted to investigate the performance of selected popular object detection models for UAV detection under challenging real-world surveillance conditions such as complex backgrounds and rainy conditions. We added Rain effects synthetically to test set images only at three different amounts (drizzle, heavy and torrential). In our framework, we have generated custom datasets covering various types of drones against a variety of environmental conditions: Sky Background Dataset (SBD), Complex Background Dataset (CBD), and Rainy Test Set (RTS). The datasets also include images of birds. The initial findings demonstrated favorable accuracy levels on the proposed datasets, particularly for the YOLOv5 and YOLOv8. However, YOLOv8 showed better results for the SBD, while YOLOv5 outperformed YOLOv8 on the CBD. Faster-RCNN could perform better with a higher number of epochs during training. However, this work is interested in examining and comparing the models' performance restricted to the same number of training epochs. The YOLOv5 achieved the highest $mAP_{50}$ among all the used models on the clean test set. However, with the rain artifacts, the YOLOv5, YOLOv8, Faster-RCNN, and RetinaNet performance degraded. The image scale variation is still an issue for YOLOv8 and YOLOv5, but FasterRCNN with backbone Resnet-101 and FPN performs better at different scales. The results and findings of this work are expected to inspire further research in more robust UAV detection, e.g., to build more challenging datasets.

Acknowledgements: This work was supported partially by SDAIA-KFUPM JRC for AI Grant No. JRC-AI-RFP-17 and by KFUPM DROC Grant No. INSS2309